\documentclass[runningheads]{llncs}
\usepackage[T1]{fontenc}
\usepackage{graphicx,verbatim}
\usepackage{multirow}
\usepackage{xcolor} 
\usepackage{amsmath}
\usepackage{booktabs}
\usepackage{subcaption}
\usepackage{amssymb}
\begin{document}
\title{SPENet: Self-guided Prototype Enhancement Network for Few-shot Medical Image Segmentation}
\titlerunning{SPENet}
\author{Chao Fan\inst{1} 
\and Xibin Jia\inst{1} \thanks{corresponding author} 
\and Anqi Xiao\inst{2}
\and Hongyuan Yu\inst{3} 
\and Zhenghan Yang\inst{4} 
\and Dawei Yang\inst{4} 
\and Hui Xu\inst{4} 
\and Yan Huang\inst{2} 
\and Liang Wang\inst{2} }

\authorrunning{C. Fan et al.}

\institute{School of Computer Science, Beijing University of Technology, Beijing, China 
\email{jiaxibin@bjut.edu.cn} \\
\and
Institute of Automation Chinese Academy of Sciences, Beijing, China \and Multimedia Department Xiaomi Inc, Beijing, China \and
Capital Medical University Affiliated Beijing Friendship Hospital, Beijing, China
}

\maketitle              % typeset the header of the contribution
\begin{abstract}
Few-Shot Medical Image Segmentation (FSMIS) aims to segment novel classes of medical objects using only a few labeled images.
Prototype-based methods have made significant progress in addressing FSMIS. However, they typically generate a single global prototype for the support image to match with the query image, overlooking intra-class variations. To address this issue, we propose a Self-guided Prototype Enhancement Network (SPENet).
Specifically, we introduce a Multi-level Prototype Generation (MPG) module, which enables multi-granularity measurement between the support and query images by simultaneously generating a global prototype and an adaptive number of local prototypes.
Additionally, we observe that not all local prototypes in the support image are beneficial for matching, especially when there are substantial discrepancies between the support and query images.
To alleviate this issue, we propose a Query-guided Local Prototype Enhancement (QLPE) module, which adaptively refines support prototypes by incorporating guidance from the query image, thus mitigating the negative effects of such discrepancies.
Extensive experiments on three public medical datasets demonstrate that SPENet outperforms existing state-of-the-art methods, achieving superior performance.

\keywords{Few-shot learning  \and Medical image segmentation \and Optimal transport.}
\end{abstract}
\section{Introduction}
Medical image segmentation aims to precisely delineate lesion or organ regions and plays a critical role in disease diagnosis and treatment planning~\cite{zhu2022multimodal,sherer2021metrics}.
Deep learning-based automatic medical image segmentation methods~\cite{ronneberger2015u,chen2021transunet,xing2024segmamba,fan2025slicemamba} have made significant progress in recent years. However, training these data-driven models demands extensive well-annotated datasets, which are particularly challenging due to privacy concerns and the need for clinical expertise. Moreover, these pre-trained models struggle to segment novel classes when only a few annotated images are available.

\begin{figure}
    \centering
    \includegraphics[width=0.82\linewidth]{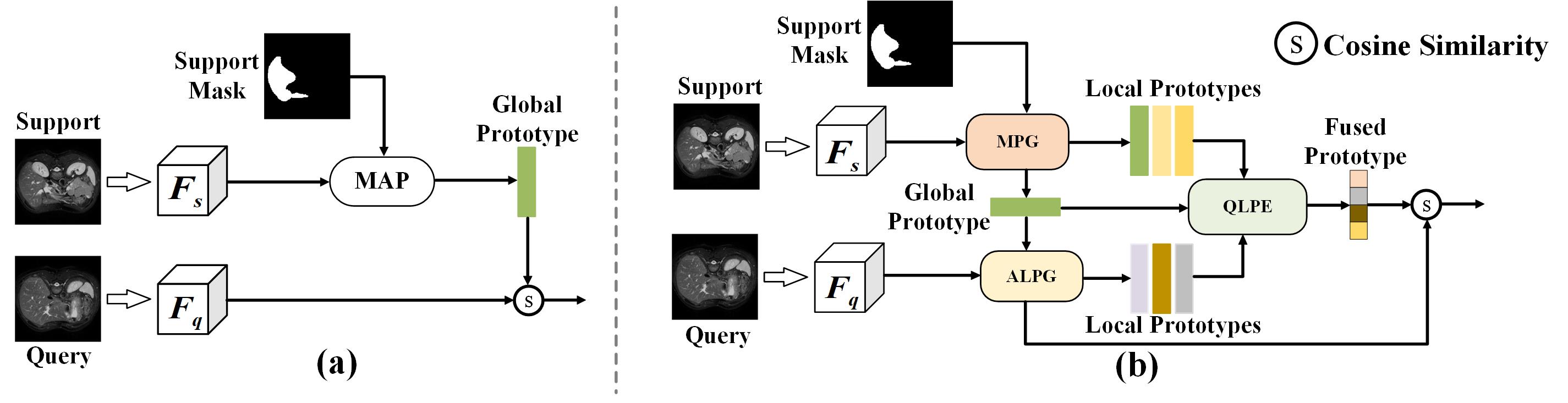}
    \caption{Comparison between (a) the classic prototype-based network and (b) our proposed Self-guided Prototype Enhancement Network (SPENet).}
    \label{fig:ab_fig}
\end{figure}

Few-shot learning has emerged as a promising solution in medical image segmentation. The core concept of FSMIS involves utilizing information from a limited number of labeled images (support set) to segment unlabeled images (query set) of the same class. 
Prototype-based methods~\cite{zhu2023few,ding2023few,tang2024few,ouyang2022self} have become mainstream in FSMIS due to their generalization ability and robustness against noise. As shown in Fig.~\ref{fig:ab_fig}(a), the classic prototype-based methods compress the support features into a global prototype using Masked Average Pooling (MAP)~\cite{zhang2020sg}, and then compute similarity with the query features. Despite its excellent performance, this scheme has notable drawbacks: the global prototype obtained via MAP loses local details, failing to effectively capture intra-class variations between the support and query images.

%Although this scheme has achieved excellent performance, it has notable drawbacks: the average pooling operation can lose local details, and the global prototype cannot effectively measure the intra-class variations (\textit{e.g.}, size, shape, appearance) between the support and query images.

Existing methods attempt to address this issue by mining finer-grained prototypes. For example, Ouyang et al.~\cite{ouyang2022self} proposed ALPNet, which divides the support image into fixed-size non-overlapping grids and extracts local prototypes from these regions. Tang et al.~\cite{tang2024few} introduced DSPNet, which employs a clustering algorithm to generate a fixed number of local prototypes for the support image and fuses them through an attention mechanism. However, these methods have limitations. ALPNet compromises the integrity of the local regions, while DSPNet overlooks the variations in the sizes of different lesions or organs, resulting in suboptimal segmentation due to the fixed number of local prototypes. 
Moreover, these methods overlook the fact that, owing to substantial intra-class variations (\textit{e.g.}, size, shape, appearance), not all local prototypes extracted from the support image contribute positively to guiding the segmentation of the query image.

To this end, we propose a Self-guided Prototype Enhancement Network named SPENet for few-shot medical image segmentation, as shown in Fig.~\ref{fig:ab_fig}(b). An ideal prototype should maintain global semantic information while preserving the local details of the object. To achieve this, we first design a Multi-level Prototype Generation (MPG) module that generates both global and local prototypes. Considering the diversity in the size of medical objects, we propose an Adaptive Local Prototype Generation (ALPG) module within MPG to generate an adaptive number of local prototypes based on the size of the object. Furthermore, to mitigate the impact of support local prototypes that differ significantly from the query image, we develop a Query-guided Local Prototype Enhancement (QLPE) module. Specifically, we utilize optimal transport to evaluate and re-weight the support local prototypes based on information from the query image. Finally, the global and refined local prototypes from the support image are fused for matching. Extensive experiments on three public medical benchmarks, including Abd-MRI, Abd-CT, and Card-MRI, demonstrate that SPENet achieves leading performance.
%

% Our contributions are summarized as follows: (1) We propose a Multi-level Prototype Generation module, which enables multi-granularity comparison between the support and query images. (2) We propose a Query-guided Local Prototype Enhancement module that effectively mitigates the interference of detrimental support local prototypes using the optimal transport algorithm. (3) Extensive experiments on three public medical benchmarks, including Abd-MRI, Abd-CT, and Card-MRI, demonstrate that SPENet achieves leading performance.

\section{Methodology}
\textbf{Problem Definition}: The goal of FSMIS is to learn a segmentation model from a training set \( D_{train} \) with classes \( C_{train} \), and then evaluate the model on a test set \( D_{test} \) with novel classes \( C_{test} \), without re-training, where \( C_{train} \cap C_{test} = \emptyset \). Following previous works~\cite{ouyang2022self}, we adopt the commonly used episode paradigm for training and testing. Specifically, we randomly sample a series of episodes from \( D_{train} \) and \( D_{test} \), resulting in $D_{train} = \{S_i,Q_i\}_{i=1}^{N_{tr}} $ and $D_{test} = \{S_i,Q_i\}_{i=1}^{N_{te}} $. Each episode consists of a support set $S=\{(I_s,M_s)\}$ and a query set $Q=\{(I_q,M_q)\}$, where \( I \) and \( M \) represent the image and its corresponding ground-truth, respectively. During training, the model predicts the segmentation mask \( M_q^{pre} \) for \( I_q \) under the supervision of \( M_q \) based on the information in \( S \). After several episodes, we obtain a trained segmentation model, which is used to predict \( I_q \) and compared with \( M_q \) to evaluate the segmentation performance.

\begin{figure*}[htp]
    \centering
    \includegraphics[width=0.9\linewidth]{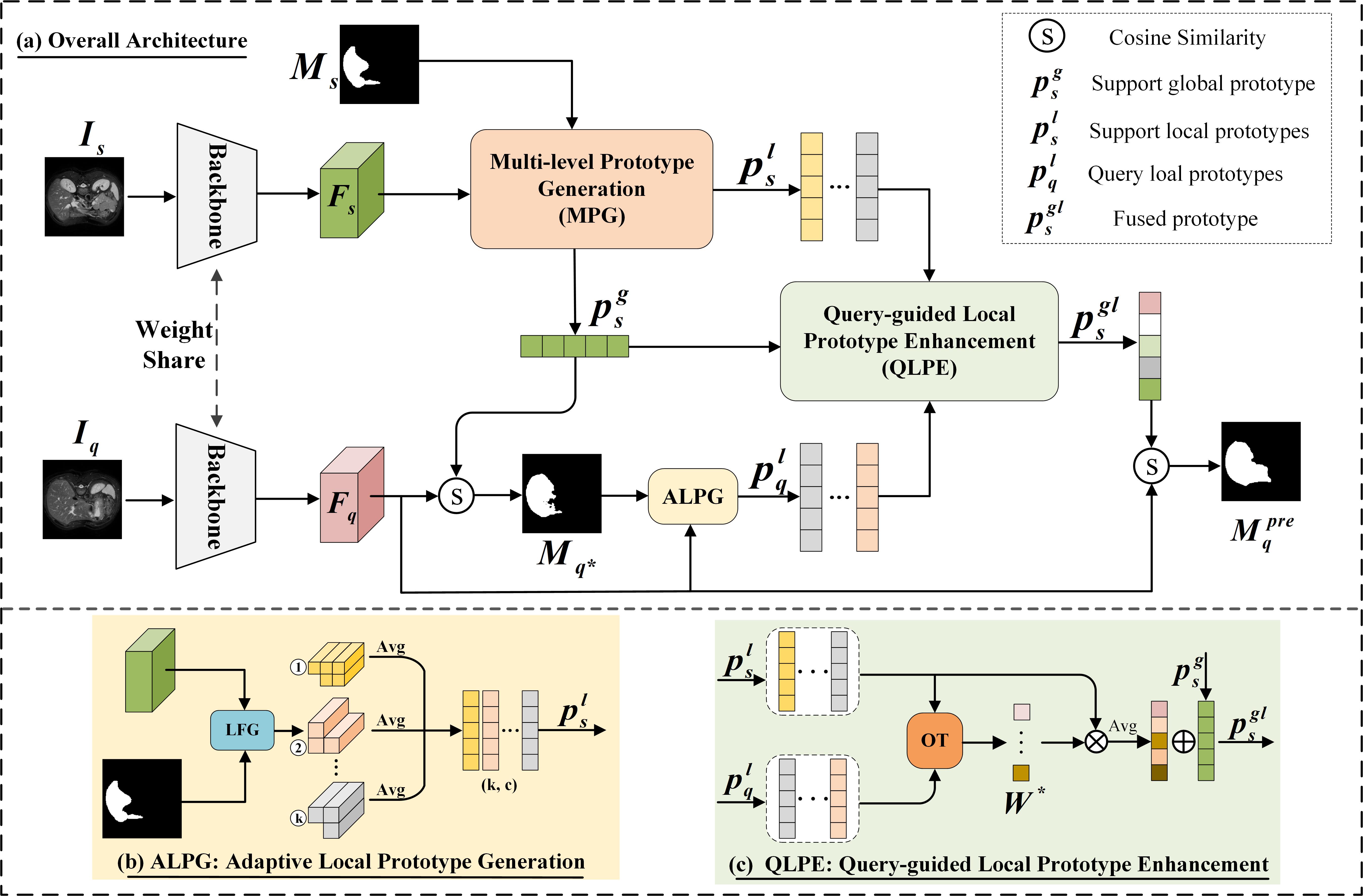}
    \caption{(a) The overview of the proposed SPENet; (b) The architecture of ALPG module; (c) The architecture of QLPE module.}
    \label{fig:archi}
\end{figure*}

\subsection{Overall Architecture}
As shown in Fig.~\ref{fig:archi}(a), our proposed SPENet consists of three main modules: \textit{(i)} a shared feature extraction network \( f_\theta(\cdot) \); \textit{(ii)} a Multi-level Prototype Generation (MPG) module; and \textit{(iii)} a Query-guided Local Prototype Enhancement (QLPE) module. Specifically, given the support image \( I_s \) and the query image \( I_q \), both are passed through the parameter-shared feature extraction network \( f_\theta(\cdot) \) to obtain their respective feature representations $F_s = f_\theta(I_s)$ and $F_q = f_\theta(I_q)$. 
In the support branch, \( F_s \) is passed into the MPG module, which consists of Masked Average Pooling (MAP) and Adaptive Local Prototype Generation (ALPG), to generate multi-level prototypes using the foreground mask $M_s$. This process produces a global prototype \( p_s^g = \text{MAP}(F_s, M_s) \) and multiple local prototypes \( p_s^l = \text{ALPG}(F_s, M_s) \), where the number of local prototypes \( p_s^l \) dynamically depends on the size of foreground in \( M_s \).
In the query branch, \( F_q \) first computes similarity with \( p_s^g \) to obtain an initial predicted mask \( M_{q*} \). The ALPG module is then applied to generate local prototype representations \( p_q^l = \text{ALPG}(F_q, M_{q*}) \) for the query image. 
Subsequently, \( p_s^l\), \( p_q^l\), and $p_s^g$ are passed into the QLPE module for prototype optimization and fusion, resulting in the final support prototype $p_s^{gl}$. Finally, the predicted mask $M_q^{pre}$ is obtained by computing the similarity between \( p_s^{gl} \) and $F_q$, \textit{i.e.}, $M_q^{pre} = sim(p_s^{gl},F_q)$, where $sim(\cdot)$ denotes cosine similarity.

\subsection{Multi-level Prototype Generation (MPG)}
Significant intra-class variation exists between support and query images~\cite{zhu2023few,wang2023rethinking}. Existing methods~\cite{ouyang2022self,zhu2023few} tackle this by mining fine-grained prototypes, but they often use a fixed number of local prototypes. These approaches overlook the diverse sizes of lesions or organs, significantly impacting segmentation performance.
To address this limitation, we propose a Multi-level Prototype Generation (MPG) module that simultaneously generates a global prototype and an adaptive number of local prototypes. The MPG module consists of two key components: the Masked Average Pooling (MAP) for the generation of the global prototype, and the Adaptive Local Prototypes Generation (ALPG) module for producing the local prototypes. Given the support features $F_s \in \mathcal{R}^{C \times h\times w}$ and the corresponding foreground mask $M_s \in \mathcal{R}^{H\times W}$, the global prototype $p_s^g \in \mathcal{R}^{C \times 1}$ via MAP can be derived as follows:
\begin{equation} \label{eq:pgs}  
p_s^g = \frac{\sum_{i,j} F_s(C, i, j) \otimes M_s(i, j)}{\sum_{i,j} M_s(i, j)}  
\end{equation}  
where $\otimes$ represents the Hadamard product and the spatial size of the feature $F_s$ will be resized to match the size of $M_s$.

Inspired by ~\cite{li2021adaptive}, we introduce the ALPG module to generate an adaptive number of local prototypes, as illustrated in Fig.~\ref{fig:archi}(b). The support feature $F_s$ and the support mask $M_s$ are processed by Local Feature Generation (\textbf{LFG}) to produce $k$ local features. 
% Average pooling is then applied to each local feature to derive corresponding local prototypes, with $k$ varying dynamically based on the size of the foreground in $M_s$. 
The LFG process begins by extracting the positions of foreground pixels from \( M_s \) as $p_{coor}$, then randomly selecting one as the initial cluster center \( S_1 \). Next, the algorithm identifies the point within the remaining $p_{coor}$ that is farthest from \( S_1 \), designating it as \( S_2 \). This process continues until a total of \( k \) center points are identified.
Subsequently, the remaining points in $p_{coor}$ are assigned to the nearest cluster centers, forming \( k \) local regions. According to these local regions, the features $F_s$ can be divided into $k$ local features.
Finally, an average operation is performed on each local feature to obtain its corresponding prototype:
\begin{equation}
    p_s^l = \left\{ \text{Avg}\left( f_i \right) \mid f_i \in \textbf{LFG}(F_s, M_s, k), \quad i = 1, 2, \dots, k \right\}
\end{equation}
where $Avg(\cdot)$ represents the average pooling operation and $\textbf{LFG}$ refers to the local feature generation operation.

The number of local prototypes \( k \) is dynamically determined by the size of the foreground in $M_s$, formulated as follows:
\begin{equation}
\label{eq:kmax}
k = \min\left(\max\left(\left\lfloor \frac{\text{sum}(M_s)}{C_s} \right\rfloor, 1\right), k_{max}\right)
\end{equation}
where $sum(M_s)$ denotes the total number of foreground pixels in $M_s$, \( C_s \) indicates the number of pixels in each local region, \( k_{\text{max}} \) is the maximum allowable number of local prototypes. Additionally, the $max(\cdot)$ function guarantees that at least one local prototype is generated for extremely small targets, ensuring all targets produce valid prototypes.

\subsection{Query-guided Local Prototypes Enhancement (QLPE)}
Existing methods directly match all local prototypes from the support image with the query features, ignoring intra-class variations such as size, shape, and appearance, which prevents achieving optimal results. For example, when the support image shows a liver tumor and the query image depicts a normal liver, not all local prototypes are beneficial.
% Existing methods directly match local prototypes with query features, ignoring the influence of anomalous local prototypes. For instance, when the support image shows a liver tumor and the query image depicts a normal liver, the local prototype representing the tumor is considered an anomalous prototype. Neglecting these anomalous prototypes can significantly affect the matching results. 
%
A straightforward solution is to calculate the cosine similarity between the local prototypes of the support and query images and then re-weight them. However, this method independently assesses the similarity between each pair of local prototypes, overlooking their relationships with others, which leads to suboptimal results. 
To address this issue, we design a Query-guided Local Prototype Enhancement module (QLPE) based on the Optimal Transport algorithm, as shown in Fig.~\ref{fig:archi}(c). 
In particular, we define (\( 1 - S \)) as the cost matrix, where $S\in \mathcal{R}^{m\times n}$ is the similarity matrix between \( p_s^l \) and \( p_q^l \), with \( m \) and \( n \) representing the number of local prototypes in \( p_s^l \) and \( p_q^l \), respectively. The transport plan $T\in \mathcal{R}^{m \times n}$ is obtained by optimizing the following function:

\begin{equation}  
\label{eq:ot}
\min_T \sum_{i=1}^{m} \sum_{j=1}^{n} T_{ij} \cdot (1-S(i, j)) + \epsilon \cdot H(T)  
\end{equation}
where $H(T) = -\sum_{i=1}^{m} \sum_{j=1}^{n} T_{ij} \log T_{ij}$ represents the entropy of the transportation plan, promoting a uniform distribution of its elements, and \( \epsilon \) is the regularization parameter, set empirically to 0.1. 

The transport matrix $T_{ij}$ satisfies the constraints \( \sum_{j=1}^{n} T_{ij} = \mu_i\), for all \(i \in \{1, 2, \ldots, m\} \) and \( \sum_{i=1}^{m} T_{ij} = \nu_j\) for all \(j \in \{1, 2, \ldots, n\}\). The \( \mu_i \) represents the importance distribution of the \( i \)-th local prototype in \( p_s^l \), and \( \nu_j \) represents the importance distribution of the \( j \)-th local prototype in \( p_q^l \). Moreover, \( \mu_i \) and \( \nu_j \) satisfy \( \sum_{i=1}^m \mu_i = 1 \) and \( \sum_{j=1}^n \nu_j = 1 \).

Using the Sinkhorn algorithm~\cite{cuturi2013lightspeed} to optimize Eq.~\ref{eq:ot} produces the optimal transport matrix $T^*$, enabling the determination of the weights for each local prototype in \( p_s^l \):
\begin{equation}
    W^* = sum(T^* \otimes S,axis=1)
\end{equation}
where \( \otimes \) represents the Hadamard product, and \( \text{axis}=1 \) indicates summation along the column dimension.

Finally, the local prototypes in \( p_s^l \) are re-weighted using \( W^* \) and averaged, then fused with the global prototype $p_s^g$ of the support image to obtain the final prototype representation \( p_s^{gl} \):
\begin{equation}
    p_s^{gl} = p_s^g + Avg(p_s^l\otimes W^*)
\end{equation}

\section{Experiments}
\subsection{Datasets and Evaluation}
We evaluate our proposed SPENet on three public medical datasets, including \textbf{Abd-MRI}~\cite{kavur2021chaos}, \textbf{Abd-CT}~\cite{landman2015miccai}, and \textbf{Card-MRI}~\cite{zhuang2018multivariate}. Abd-MRI is from the 2019 ISBI Combined Healthy Abdominal Organ Segmentation (CHAOS) challenge, consisting of 20 abdominal 3D MRI scans. Abd-CT is from the 2015 MICCAI Multi-Atlas Abdomen Labeling challenge, comprising 30 abdominal 3D CT scans. Card-MRI is from the 2019 MICCAI Automatic Cardiac Chamber and Myocardium Segmentation challenge, consisting of 35 clinical 3D cardiac MRI scans.

We use the widely adopted Dice Similarity Coefficient (\textbf{DSC}) to evaluate the performance of the segmentation model. Following previous work~\cite{ouyang2022self}, we employ two supervision settings in the experiment. In \textbf{Setting I}, test classes may occur in the background region of training slices. For example, the liver (training class) and spleen (test class) may appear in the same slice. \textbf{Setting II} is stricter and requires training and test classes to be entirely separate within slices. Here, slices containing test classes are excluded during training, ensuring the model does not encounter test classes beforehand. However, Setting II is not feasible for the Card-MRI dataset, where all classes co-occur in one slice.

\begin{table}[!ht]
\centering
\caption{Comparison results in DSC (\%) on Abd-CT, Abd-MRI, and Card-MRI datasets. The best results are highlighted in bold.}
\scalebox{0.81}{
\renewcommand{\arraystretch}{1.2} 
\setlength{\tabcolsep}{0.5pt}
\begin{tabular}{l|c|ccccc|ccccc|cccc}
\hline

\multicolumn{2}{c|}{\multirow{2}{*}{Methods}} & \multicolumn{5}{c|}{\textbf{Abd-MRI}}       
& \multicolumn{5}{c|}{\textbf{Abd-CT}}   & \multicolumn{4}{c}{\textbf{Card-MRI}}\\ 
\cline{3-16} 
\multicolumn{2}{c|}{}  
& LK    & RK   & Spleen &Liver & \textbf{Mean} & LK   & RK    & Spleen & Liver & \textbf{Mean}
& BP   & MYO    & RV & \textbf{Mean}
\\ \hline
\multirow{8}{*}{\rotatebox{90}{Setting I}}     
& ALPNet%~\cite{ouyang2022self}   
& \underline{81.92}   & 85.18   & 72.18  & 76.10 &78.84   
& 72.36  & 71.81  & 70.96 & 78.29  & 73.35 
& 83.99  & 66.74 & 79.96  & 76.90\\
& Q-Net%~\cite{shen2023q}   
& 68.36      & 84.41  & \underline{76.69} & 73.54   & 68.65 
& 69.39      & 55.63    & 56.82    & 68.65     & 62.63
& 89.15    & 64.52    & 78.19     & 77.28\\
& CRAPNet%~\cite{ding2023few} 
& 80.38     & 86.42  & 74.32  & 76.46   & 79.39        
& 74.69       & 74.18    & 70.37   & 75.41   & 77.66 
& 83.02    & 65.48   & 78.27   & 75.59 \\
& RPT%~\cite{zhu2023few}    
& 80.72   & \textbf{89.82}  & 76.37  & \underline{82.86}  & \underline{82.44}    & 77.05    & \underline{79.13}   & \underline{72.58}  & \underline{82.57}          & \underline{77.83}   
& \underline{89.90}    & \underline{66.91}   & \underline{80.78}   & \underline{79.19}\\
& PAMI%~\cite{zhu2024partition}   
& 81.38     & 88.43   & 76.37    & 82.59  & 82.38        
& 76.52     & \textbf{80.57}   & 72.38    & 81.32 & 77.69
 & 89.57    & 66.82 & 80.17 & 78.85\\
& DSPNet%~\cite{tang2024few}    
& 81.88          & 85.37         & 70.93          & 75.06         & 78.31        
& \underline{78.01}         & 74.54         & 69.31          & 69.32       & 72.79
& 87.75         & 64.91          & 79.73       & 77.46 \\
& \textbf{Ours}    
& \textbf{81.98}   & \underline{89.18}  & \textbf{80.41}  &\textbf{83.02}    & \textbf{83.65}            
& \textbf{80.01}    & 73.39   & \textbf{80.74}  &\textbf{82.76}    &\textbf{79.23} 
&\textbf{90.15}         & \textbf{69.99}          & \textbf{81.80}       & \textbf{80.64}\\ \hline
\multirow{8}{*}{\rotatebox{90}{Setting II}} 
& ALPNet %~\cite{ouyang2022self}  
& 73.63     & 78.39   & 67.02   & 73.05    & 73.02  
& 63.34       & 54.82      & 60.25   & 73.65   & 63.02 
& -      &-   & -   & -
\\
& Q-Net%~\cite{shen2023q}                                            
& 73.96   & 81.07   & 65.39   & 72.36   & 73.20        
& 66.25    & 62.36  & 67.35 & \underline{77.33}   & 68.32         
& -      &-   & -   & -\\
& CRAPNet%~\cite{ding2023few}  
& 74.66          & 82.77          & 70.82           & 73.82          & 75.52         
& 70.91& 67.33          & 70.17           & 70.45    & 69.72  
& -      &-   & -   & -\\
& RPT%~\cite{zhu2023few}         
& \textbf{78.33}    & 86.01    & 75.46  & 76.37          & 79.04        
& \underline{72.99}   & \underline{67.73}     & 70.80  & 75.24 &\underline{71.69}      
& -      &-   & -   & -\\
& PAMI%~\cite{zhu2024partition}                                                  
& 74.51     &\underline{86.73}   & \underline{75.80}   &\textbf{81.09} &\underline{79.53} & 72.36          & 67.54          & \underline{71.95}           & 74.13          & 71.49
& -      &-   & -   & -\\
& DSPNet%~\cite{tang2024few}                                                
& 76.47          & 82.01          & 68.27           & 78.56          & 76.33         
& 68.46          & 63.55          & 66.48           & 69.16          & 66.17         
& -      &-   & -   & -\\
& \textbf{Ours}     
& \underline{76.94}  & \textbf{88.07}   & \textbf{76.78}  & \underline{80.88}   & \textbf{80.67}            
& \textbf{77.40}       & \textbf{72.21}        & \textbf{81.07}      & \textbf{84.42}     & \textbf{78.77}          
& -      &-   & -   & -\\ \hline

\end{tabular}}

\label{tababd_data}
\end{table}

\subsection{Implementation Details}
To ensure methodological consistency and reproducibility, we adopt the following implementation protocol. The backbone architecture employs a ResNet-101~\cite{he2016deep} network initialized with weights pre-trained on the MS-COCO dataset~\cite{lin2014microsoft}, serving as our foundational feature extractor. Following previous work~\cite{ouyang2022self}, all 3D medical scans are reformatted into 2D axial slices and resized to $256 \times 256$. The proposed model is trained using an SGD optimizer with an initial learning rate of 1e-3, a weight decay of 0.0005, and a momentum coefficient of 0.9. Consistent with most current FSMIS methodologies~\cite{ouyang2022self,zhu2023few}, we adopt a 1-way-1-shot learning protocol, set the batch size to 1, and train the model for 30$k$ iterations. We set $k_{max}$ and $C_s$ in Eq.~\ref{eq:kmax} to 24 and 50, respectively. To validate the efficacy and generalizability of the proposed model, we perform 5-fold cross-validation and report the mean segmentation result across all folds. The implementation is carried out on a computational platform running Ubuntu 22.04, with Python version 3.8.19, CUDA version 11.8, and powered by an RTX 4090 GPU, ensuring efficient training and evaluation.

\subsection{Comparison with State-of-the-art Methods}
We compare the proposed SPENet with state-of-the-art methods, including ALPNet~\cite{ouyang2022self}, Q-Net~\cite{shen2023q},  CRAPNet~\cite{ding2023few}, RPT~\cite{zhu2023few}, PAMI~\cite{zhu2024partition}, and DSPNet~\cite{tang2024few}.
\textbf{Abd-MRI and Abd-CT Datasets:} The quantitative results presented in Tab.~\ref{tababd_data} demonstrate the superior performance of SPENet on both the Abd-MRI and Abd-CT datasets under setting I and setting II. To be specific, on the Abd-MRI dataset, SPENet achieves mean DSC scores of 83.65\% and 80.67\% under setting I and setting II, respectively, surpassing RPT by 1.21\% under setting I and PAMI by 1.14\% under setting II. On the Abd-CT dataset, SPENet outperforms the second-place method, RPT, by margins of 1.40\% and 7.08\% under setting I and setting II, respectively.
\textbf{Card-MRI Dataset:} Since all classes are present in one slice of the Card-MRI dataset, only setting I is applicable. SPENet achieves the best results on the Card-MRI dataset. Specifically, our method achieves a mean DSC of 80.64\%, outperforming RPT and PAMI by 1.45\% and 1.79\%, respectively. 

\subsection{Ablation Study}
\begin{figure}[!h]
    \centering

    \begin{minipage}{0.45\textwidth}
        \centering
        \captionsetup{type=table} 
        \caption{Ablation results of key components on the Abd-MRI dataset.}
        \resizebox{\textwidth}{!}{
        \renewcommand{\arraystretch}{1}
        \begin{tabular}{@{}cccccc@{}}
            \toprule
            Baseline        & MPG*  & MPG     &QLPE & DSC  & Improve \\
            \midrule
            \checkmark    &  &   &  & 80.12   & -\\
            \checkmark   &\checkmark  &   &   & 81.85  & \textcolor{red}{+1.73}\\
            \checkmark    &   & \checkmark  &   & 82.59  &\textcolor{red}{+2.47}\\
            \checkmark    &  &\checkmark  &\checkmark &\textbf{83.65} & \textcolor{red}{+3.53}\\
            \bottomrule
        \end{tabular}
        }
        \label{ablation_table}
    \end{minipage}%
    \hfill
    % 右侧图像
    \begin{minipage}{0.45\textwidth}
        \centering
        \includegraphics[width=\textwidth]{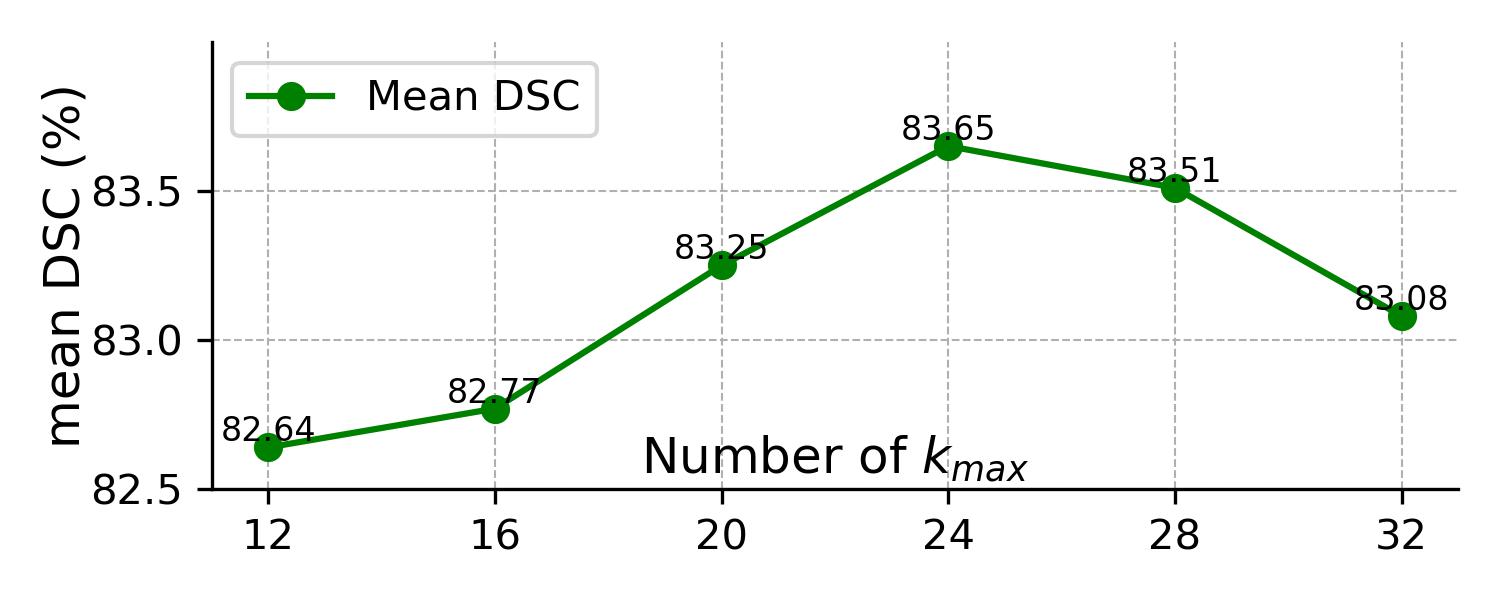} 
        \caption{Analysis of the maximum number of local prototypes.}
        \label{fig:hyper}
    \end{minipage}

\end{figure}

We perform a series of ablation studies on Abd-MRI under setting I to verify the effectiveness of each component in SPENet, as shown in Tab.~\ref{ablation_table}. 
\textbf{Baseline} refers to the model without the MPG and QLPE modules, while keeping all other training parameters.
\textbf{$\textbf{MPG}^*$} denotes the MPG module that generates a global prototype and a \textbf{fixed number} of local prototypes, resulting in a 1.73\% improvement in mean DSC when integrated into the baseline. 
\textbf{MPG} can generate an \textbf{adaptive number} of local prototypes, which improves the mean DSC by 2.47\% when added to the baseline. 
When the baseline is equipped with both the \textbf{MPG} and \textbf{QLPE} modules, it achieves a 3.53\% increase in mean DSC over the baseline. These results confirm the effectiveness of the key modules in our proposed SPENet.
\noindent\textbf{Hyperpameter Evaluation}: We also conduct experiments with different $k_{max}$ in Eq.~\ref{eq:kmax} to explore the impact of the number of local prototypes on model performance. Fig.~\ref{fig:hyper} shows that model performance improves with increasing $k_{max}$ when \(k_{\text{max}} < 24\), but slightly declines thereafter while still outperforming other methods. We believe this phenomenon occurs because dividing into too many local regions can compromise semantic information, leading to inconsistent matching~\cite{zhang2021few}. Therefore, we set $k_{max}=24$ in the experiment.

\section{Conclusion}
In this paper, we introduce the Self-Guided Prototype Enhancement Network (SPENet) to address the challenge of intra-class variation in FSMIS. SPENet comprises two key modules: the Multi-level Prototype Generation (MPG) module and the Query-guided Local Prototype Enhancement (QLPE) module. 
The MPG module enables multi-granularity measurement by simultaneously generating a global prototype and an adaptive number of local prototypes. The global prototype preserves semantic information, while the local prototypes effectively capture subtle variations between the support and query images.
Complementing this, the QLPE module aims to mitigate the interference of detrimental local prototypes in the support image through an optimal transport algorithm, thereby further reducing intra-class variation.
Extensive experiments conducted on three public medical image segmentation datasets demonstrate the significant advantages of SPENet, highlighting its superior performance.

\begin{credits}
\subsubsection{\ackname} This work is partly supported by National Natural Science Foundation of China No.62476015, 62171298, 82071876, 82372043, 82371904.

\subsubsection{\discintname}
The authors have no competing interests to declare that are relevant to the content of this article.
\end{credits}
\bibliographystyle{splncs04}
\bibliography{Paper-0184}

\begin{thebibliography}{10}
\providecommand{\url}[1]{\texttt{#1}}
\providecommand{\urlprefix}{URL }
\providecommand{\doi}[1]{https://doi.org/#1}

\bibitem{chen2021transunet}
Chen, J., Lu, Y., Yu, Q., Luo, X., Adeli, E., Wang, Y., Lu, L., Yuille, A.L.,
  Zhou, Y.: Transunet: Transformers make strong encoders for medical image
  segmentation. arXiv preprint arXiv:2102.04306  (2021)

\bibitem{cuturi2013lightspeed}
Cuturi, M.: Lightspeed computation of optimal transportation distances.
  Advances in Neural Information Processing Systems  \textbf{26}(2),
  2292--2300 (2013)

\bibitem{ding2023few}
Ding, H., Sun, C., Tang, H., Cai, D., Yan, Y.: Few-shot medical image
  segmentation with cycle-resemblance attention. In: Proceedings of the
  IEEE/CVF Winter Conference on Applications of Computer Vision. pp. 2488--2497
  (2023)

\bibitem{fan2025slicemamba}
Fan, C., Yu, H., Huang, Y., Wang, L., Yang, Z., Jia, X.: Slicemamba with neural
  architecture search for medical image segmentation. IEEE Journal of
  Biomedical and Health Informatics  (2025)

\bibitem{he2016deep}
He, K., Zhang, X., Ren, S., Sun, J.: Deep residual learning for image
  recognition. In: Proceedings of the IEEE conference on computer vision and
  pattern recognition. pp. 770--778 (2016)

\bibitem{kavur2021chaos}
Kavur, A.E., Gezer, N.S., Bar{\i}{\c{s}}, M., Aslan, S., Conze, P.H., Groza,
  V., Pham, D.D., Chatterjee, S., Ernst, P., {\"O}zkan, S., et~al.: Chaos
  challenge-combined (ct-mr) healthy abdominal organ segmentation. Medical
  Image Analysis  \textbf{69},  101950 (2021)

\bibitem{landman2015miccai}
Landman, B., Xu, Z., Igelsias, J., Styner, M., Langerak, T., Klein, A.: Miccai
  multi-atlas labeling beyond the cranial vault--workshop and challenge. In:
  Proc. MICCAI Multi-Atlas Labeling Beyond Cranial Vault—Workshop Challenge.
  vol.~5, p.~12 (2015)

\bibitem{li2021adaptive}
Li, G., Jampani, V., Sevilla-Lara, L., Sun, D., Kim, J., Kim, J.: Adaptive
  prototype learning and allocation for few-shot segmentation. In: Proceedings
  of the IEEE/CVF conference on computer vision and pattern recognition. pp.
  8334--8343 (2021)

\bibitem{lin2014microsoft}
Lin, T.Y., Maire, M., Belongie, S., Hays, J., Perona, P., Ramanan, D.,
  Doll{\'a}r, P., Zitnick, C.L.: Microsoft coco: Common objects in context. In:
  Computer Vision--ECCV 2014: 13th European Conference, Zurich, Switzerland,
  September 6-12, 2014, Proceedings, Part V 13. pp. 740--755. Springer (2014)

\bibitem{ouyang2022self}
Ouyang, C., Biffi, C., Chen, C., Kart, T., Qiu, H., Rueckert, D.:
  Self-supervised learning for few-shot medical image segmentation. IEEE
  Transactions on Medical Imaging  \textbf{41}(7),  1837--1848 (2022)

\bibitem{ronneberger2015u}
Ronneberger, O., Fischer, P., Brox, T.: U-net: Convolutional networks for
  biomedical image segmentation. In: Medical image computing and
  computer-assisted intervention--MICCAI 2015: 18th international conference,
  Munich, Germany, October 5-9, 2015, proceedings, part III 18. pp. 234--241.
  Springer (2015)

\bibitem{shen2023q}
Shen, Q., Li, Y., Jin, J., Liu, B.: Q-net: Query-informed few-shot medical
  image segmentation. In: Proceedings of SAI Intelligent Systems Conference.
  pp. 610--628. Springer (2023)

\bibitem{sherer2021metrics}
Sherer, M.V., Lin, D., Elguindi, S., Duke, S., Tan, L.T., Cacicedo, J., Dahele,
  M., Gillespie, E.F.: Metrics to evaluate the performance of auto-segmentation
  for radiation treatment planning: A critical review. Radiotherapy and
  Oncology  \textbf{160},  185--191 (2021)

\bibitem{tang2024few}
Tang, S., Yan, S., Qi, X., Gao, J., Ye, M., Zhang, J., Zhu, X.: Few-shot
  medical image segmentation with high-fidelity prototypes. Medical Image
  Analysis p. 103412 (2024)

\bibitem{wang2023rethinking}
Wang, Y., Sun, R., Zhang, T.: Rethinking the correlation in few-shot
  segmentation: A buoys view. In: Proceedings of the IEEE/CVF Conference on
  Computer Vision and Pattern Recognition. pp. 7183--7192 (2023)

\bibitem{xing2024segmamba}
Xing, Z., Ye, T., Yang, Y., Liu, G., Zhu, L.: Segmamba: Long-range sequential
  modeling mamba for 3d medical image segmentation. In: International
  Conference on Medical Image Computing and Computer-Assisted Intervention. pp.
  578--588. Springer (2024)

\bibitem{zhang2021few}
Zhang, G., Kang, G., Yang, Y., Wei, Y.: Few-shot segmentation via
  cycle-consistent transformer. Advances in Neural Information Processing
  Systems  \textbf{34},  21984--21996 (2021)

\bibitem{zhang2020sg}
Zhang, X., Wei, Y., Yang, Y., Huang, T.S.: Sg-one: Similarity guidance network
  for one-shot semantic segmentation. IEEE transactions on cybernetics
  \textbf{50}(9),  3855--3865 (2020)

\bibitem{zhu2022multimodal}
Zhu, Q., Wang, H., Xu, B., Zhang, Z., Shao, W., Zhang, D.: Multimodal triplet
  attention network for brain disease diagnosis. IEEE Transactions on Medical
  Imaging  \textbf{41}(12),  3884--3894 (2022)

\bibitem{zhu2023few}
Zhu, Y., Wang, S., Xin, T., Zhang, H.: Few-shot medical image segmentation via
  a region-enhanced prototypical transformer. In: International Conference on
  Medical Image Computing and Computer-Assisted Intervention. pp. 271--280.
  Springer (2023)

\bibitem{zhu2024partition}
Zhu, Y., Wang, S., Xin, T., Zhang, Z., Zhang, H.: Partition-a-medical-image:
  Extracting multiple representative sub-regions for few-shot medical image
  segmentation. IEEE Transactions on Instrumentation and Measurement  (2024)

\bibitem{zhuang2018multivariate}
Zhuang, X.: Multivariate mixture model for myocardial segmentation combining
  multi-source images. IEEE transactions on pattern analysis and machine
  intelligence  \textbf{41}(12),  2933--2946 (2018)

\end{thebibliography}




% LNCS Homepage, \url{http://www.springer.com/lncs}, last accessed 2023/10/25
% \end{thebibliography}
\end{document}